\documentclass[letterpaper, 10 pt, journal, twoside]{IEEEtran}

\ifCLASSINFOpdf
  \usepackage[pdftex]{graphicx}
  \usepackage[export]{adjustbox}
  \usepackage[hidelinks]{hyperref}
  \hypersetup{pdfauthor=author}
  \usepackage{amsmath,amsthm,amsfonts,amssymb}
  \usepackage{cuted} 
  \usepackage{enumitem}
  \usepackage{balance}
  \usepackage{subcaption}
  \usepackage{cite}
  \usepackage{soul}
  \usepackage{multirow}
  \usepackage{orcidlink}

  \usepackage{algorithm}
  \usepackage{algpseudocode}

  \usepackage{array}
  \newcommand{\PreserveBackslash}[1]{\let\temp=\\#1\let\\=\temp}
  \newcolumntype{C}[1]{>{\PreserveBackslash\centering}m{#1}}

  \usepackage[utf8]{inputenc}
  \usepackage[T5]{fontenc}
\else

  \usepackage[dvips]{graphicx}
\fi

\begin{document}
\title{Design of a Bio-Inspired Miniature Submarine for Low-Cost Water Quality Monitoring}

\author{Quang Huy Vu$^{1}$, Quan Le$^{2}$, Manh Duong Phung$^{3}$

\thanks{$^{1}$Quang Huy Vu is with HUS High School for Gifted Students, VNU University of Science, Hanoi, 100000, Vietnam.}

\thanks{$^{2}$Quan Le is with Fulbright University Vietnam, Ho Chi Minh City, 700000, Vietnam.}

\thanks{$^{3}$Manh Duong Phung is with College of Engineering and Computer Science, VinUniversity, Hanoi, 100000, Vietnam. {\tt\footnotesize duong.pm@vinuni.edu.vn}}
}



\maketitle

\begin{abstract}
Water quality monitoring is essential for protecting aquatic ecosystems and detecting environmental pollution. This paper presents the design and experimental validation of a bio-inspired miniature submarine for low-cost water quality monitoring. Inspired by the jet propulsion mechanism of squids, the proposed system employs pump-driven water jets for propulsion and steering, combined with a pump-based buoyancy control mechanism that enables both depth regulation and water sampling. The vehicle integrates low-cost, commercially available components including an ESP32 microcontroller, IMU, pressure sensor, GPS receiver, and LoRa communication module. The complete system can be constructed at a hardware cost of approximately \$122.5, making it suitable for educational and environmental monitoring applications. Experimental validation was conducted through pool tests and field trials in a lake. During a $360^\circ$ rotation test, roll and pitch deviations remained within $\pm2^\circ$ and $\pm1.5^\circ$, respectively, demonstrating stable attitude control. Steering experiments showed a heading step response with approximately 2\,s rise time and 5\,s settling time. Depth control experiments achieved a target depth of 2.5\,m with steady-state error within $\pm0.1$\,m. Field experiments further demonstrated reliable navigation and successful water sampling operations. The results confirm that the proposed platform provides a compact, stable, and cost-effective solution for small-scale aquatic environmental monitoring.
\end{abstract}


\section{Introduction}

Water quality monitoring is essential for protecting aquatic ecosystems, ensuring public health, and supporting sustainable management of water resources \cite{chapman2022role}. Lakes, rivers, and coastal environments are increasingly threatened by pollution, eutrophication, and urban runoff. Regular monitoring of physical, chemical, and biological parameters is therefore necessary to detect contamination, evaluate ecosystem health, and guide environmental policy. However, conventional monitoring methods typically rely on manual sampling from boats or fixed stations, which can be labor-intensive, costly, and limited in spatial and temporal coverage \cite{rand2022human,lopez2023wireless}.

Recent advances in robotics and embedded systems have enabled the development of autonomous and remotely operated vehicles for environmental monitoring. Underwater robots provide the ability to perform in-situ sensing and sampling at different depths and locations while reducing human effort and operational risks \cite{petillot2019underwater, cai2023cooperative}. Despite these advantages, many existing underwater vehicles are designed for industrial or scientific missions and often involve complex mechanical systems, expensive sensors, and high operational costs \cite{petillot2019underwater,1240276}. Such systems are therefore not always suitable for routine environmental monitoring or educational and research applications with limited budgets.

Bio-inspired robotic design offers an alternative approach for improving mobility and efficiency in aquatic environments. By mimicking the locomotion strategies of marine organisms, bio-inspired underwater vehicles can achieve stable and energy-efficient motion while maintaining relatively simple mechanical structures. Several studies have explored fish-inspired propulsion, squid-like jet propulsion, and other biologically inspired mechanisms to enhance maneuverability and reduce energy consumption in underwater robots \cite{costa2018design, park2014design,xiong2023bio}.

In this work, we present the design and experimental validation of a bio-inspired miniature submarine intended for low-cost water quality monitoring. The proposed platform integrates a compact propulsion system, differential steering, and a pump-based buoyancy mechanism that enables both depth regulation and water sampling. The system is built using low-cost, commercially available components, including an ESP32 microcontroller, inertial and pressure sensors, GPS positioning, and long-range LoRa communication. This design aims to provide an affordable and easily reproducible platform suitable for environmental monitoring, research, and educational purposes.

The main contributions of this paper are threefold: (1) the design of a squid-inspired miniature submarine that integrates jet propulsion and buoyancy control to achieve four degrees of freedom (4-DoF) motion; (2) the implementation of the platform with onboard sensing and wireless communication for remote monitoring, forming a low-cost monitoring system with an estimated cost of about \$122.5; and (3) experimental validation through controlled pool experiments and field trials conducted in a lake. The results show that the proposed platform can maintain stable motion, perform controlled navigation and depth adjustment, and successfully carry out water sampling tasks in real-world environments.

The remainder of this paper is organized as follows. Section II reviews related work. Section III presents the design of the proposed miniature submarine. Section IV describes the system architecture and control. Section V reports the experimental setup and results from pool and field tests. Finally, Section VI concludes the paper and discusses future work.

\section{Related work}

Underwater robotic vehicles have been widely developed for oceanographic exploration, inspection, and environmental monitoring. Conventional underwater vehicles, including autonomous underwater vehicles (AUVs) and remotely operated vehicles (ROVs), typically rely on propeller-based propulsion systems and rigid mechanical structures. Comprehensive surveys of underwater robotic systems highlight their applications in marine research, offshore inspection, and environmental data collection \cite{neira2021review,10477402}. Although these platforms provide reliable propulsion and navigation capabilities, their relatively high cost and mechanical complexity often limit their use in small-scale environmental monitoring tasks.

To improve maneuverability and energy efficiency in aquatic environments, many researchers have explored bio-inspired underwater robots that mimic the locomotion of marine animals. Early work such as the RoboTuna project demonstrated that fish-inspired propulsion can achieve high hydrodynamic efficiency and smooth maneuverability \cite{tolkoff1999robotics}. Subsequent studies further investigated robotic fish designs that replicate oscillatory tail propulsion and body undulation \cite{li2022comprehensive, dabiri2009optimal,raj2016fish}. These systems can achieve improved agility, reduced hydrodynamic noise, and better interaction with complex environments compared with conventional propeller-driven vehicles. More recent reviews summarize the development of biomimetic underwater robots and their advantages in efficiency, maneuverability, and environmental compatibility \cite{yan2024recent,wang2020development}. Bio-inspired mechanisms have also been applied to soft underwater robots that mimic the motion of jellyfish or cephalopods to improve propulsion efficiency and adaptability \cite{4623834,shen2017biomimetic, marchese2014autonomous,bao2023review}.

Underwater robots have also been increasingly used for environmental monitoring and water quality assessment. Autonomous platforms equipped with environmental sensors can perform in-situ measurements of parameters such as temperature, turbidity, pH, dissolved oxygen, and conductivity \cite{manjakkal2021connected}. Robotic monitoring systems provide advantages over traditional manual sampling by enabling continuous data collection, improved spatial coverage, and reduced human effort. For example, an autonomous surface vehicle (ASV) designed to navigate complex inland water bodies while measuring various water quality parameters and greenhouse gas emissions is presented in \cite{dunbabin2009autonomous}. Several studies have also proposed low-cost robotic platforms for water quality monitoring using embedded sensors and wireless communication technologies \cite{adu2017water,madeo2020low}. These systems demonstrate the potential of small robotic platforms for real-time environmental monitoring in lakes and rivers.

Despite these advances, many existing underwater robots remain expensive and mechanically complex, which limits their accessibility for routine monitoring or educational applications. While some low-cost AUV platforms have been proposed, many designs still require specialized propulsion systems, expensive sensors, or sophisticated structures \cite{LI2022109056,202406956}. Consequently, there remains a need for compact and affordable underwater robotic platforms that integrate efficient propulsion, simple buoyancy control, and environmental sensing capabilities. 

In this work, we present a bio-inspired miniature submarine designed specifically for low-cost water quality monitoring. The proposed platform combines a simplified propulsion configuration, differential steering, and a pump-based buoyancy system with integrated sensing and wireless communication. The design aims to provide an accessible and reproducible robotic platform suitable for environmental monitoring, research, and educational applications.

\section{Proposed design}
The design of the proposed miniature submarine is inspired by squid locomotion and is described as follows.

\begin{figure}[ht]
\centering
\includegraphics[width=\columnwidth]{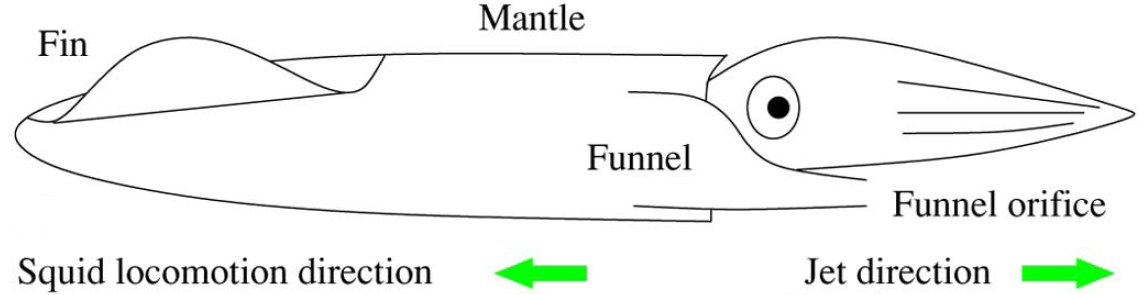}
\caption{Locomotion mechanism of a squid based on jet propulsion \cite{4623834}.}
\label{fig:squid}
\end{figure}

\subsection{Design principle}

The locomotion mechanism of a squid is based on jet propulsion which enables rapid and efficient movement in water. As illustrated in Fig. \ref{fig:squid}, the squid draws water into its mantle cavity and then forcefully expels it through a narrow funnel. When the mantle muscles contract, the water is pushed out through the funnel, generating a reaction force that propels the squid in the opposite direction according to Newton’s third law. By adjusting the orientation of the funnel, the squid can control both the direction and speed of movement. This natural propulsion strategy inspired the design of our miniature submarine. The system similarly uses a pump-driven jet mechanism to intake and expel water, which produces thrust for underwater locomotion while maintaining a compact and efficient mechanical structure.

\begin{figure*}[ht]
\centering
\includegraphics[width=0.85\textwidth]{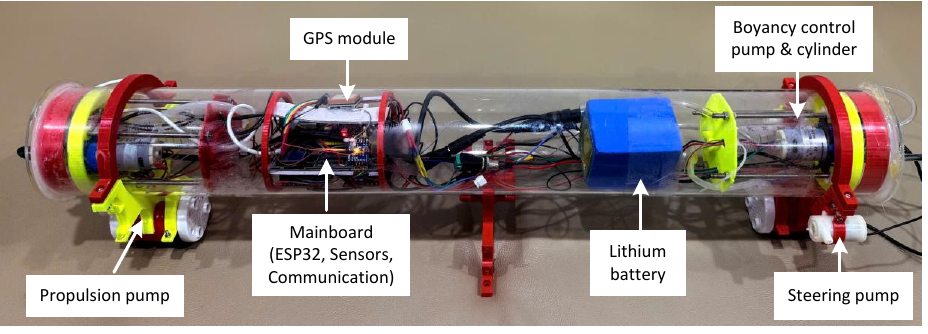}
\caption{Prototype of the squid-inspired miniature submarine with labeled main components}
\label{fig:submarine}
\end{figure*}

The proposed miniature submarine is designed as a compact cylindrical platform capable of four degrees of freedom (4-DOF) motion in underwater environments. As shown in Fig. \ref{fig:submarine}, the system integrates the main mechanical and electronic modules inside a transparent waterproof tube. Two propulsion pumps are mounted at the two ends of the vehicle, generating jet thrust that enables forward and backward movement and contributes to attitude stabilization. For directional control, two steering pumps are installed on the left and right sides near the rear end, producing lateral water jets that allow the vehicle to turn and adjust its heading. Vertical motion is achieved through a buoyancy control system consisting of two cylinders driven by DC motors, with one cylinder installed at each end of the submarine. The motors actuate the cylinders to intake or release water, thereby adjusting the internal water volume and controlling the vehicle’s buoyancy. By increasing or decreasing the amount of water inside the cylinders, the submarine can ascend or descend in the water column. In addition, the difference in water volume between the front and rear cylinders enables control of the pitch angle, allowing the vehicle to tilt upward or downward during operation. This pump-based configuration provides a compact and efficient mechanism for maneuverable underwater locomotion.

\subsection{Motion model}

To describe the motion of the proposed miniature submarine, a simplified decoupled dynamic model is adopted. The vehicle is actuated by two propulsion pumps located at the front and rear, two steering pumps mounted on the left and right sides near the rear end, and a buoyancy control mechanism consisting of two DC-motor–driven cylinders installed at the two ends of the vehicle. The propulsion pumps generate jet thrust for forward and backward motion, while the differential action of the two steering pumps produces lateral jet forces that control the vehicle’s yaw motion. Vertical movement is achieved by adjusting buoyancy through the intake or release of water in the two cylinders, and pitch is controlled by creating a volume difference between them. Assuming small roll and pitch angles and weak coupling between motion axes, the system dynamics can be approximated by four independent equations representing surge, yaw, heave, and pitch motions.

\begin{equation}
m_u \dot{u} = k_p (\omega_{p1} + \omega_{p2}) - d_u u
\end{equation}

\begin{equation}
I_z \dot{r} = k_s (\omega_{sR} - \omega_{sL}) - d_r r
\end{equation}

\begin{equation}
m_w \dot{w} = \rho g (\Delta V_1 + \Delta V_2) - d_w w
\end{equation}

\begin{equation}
I_y \dot{q} = \rho g l_b (\Delta V_1 - \Delta V_2) - d_q q - k_\theta \theta
\end{equation}

where $u$, $r$, $w$, and $q$ denote the surge velocity, yaw rate, heave velocity, and pitch rate, respectively, while $\theta$ represents the pitch angle. The control inputs $\omega_{p1}$ and $\omega_{p2}$ represent the rotational speeds of the front and rear propulsion pumps, while $\omega_{sL}$ and $\omega_{sR}$ denote the speeds of the left and right steering pumps. The variables $\Delta V_1$ and $\Delta V_2$ represent the volume changes of water in the front and rear cylinders, respectively. The parameters $m_u$ and $m_w$ represent the effective mass in the surge and heave directions, $I_z$ and $I_y$ are the yaw and pitch moments of inertia, and $d_u$, $d_r$, $d_w$, and $d_q$ are hydrodynamic damping coefficients. The parameter $l_b$ denotes the longitudinal distance from the center of mass to the buoyancy cylinders, and $k_\theta$ represents the hydrostatic pitch restoring coefficient. The constants $k_p$ and $k_s$ denote actuator gain coefficients, while $\rho$ and $g$ represent the water density and gravitational acceleration, respectively.

Based on this model, forward and backward motion are produced by the combined thrust of the two propulsion pumps, left and right turning are generated by the differential thrust of the left and right steering pumps. Vertical movement (ascending or descending) is achieved by controlling the total displaced water volume $(\Delta V_1 + \Delta V_2)$, while the pitch attitude is adjusted by controlling the differential volume $(\Delta V_1 - \Delta V_2)$ between the two cylinders.

\begin{figure*}[ht]
\centering
\includegraphics[width=0.75\textwidth]{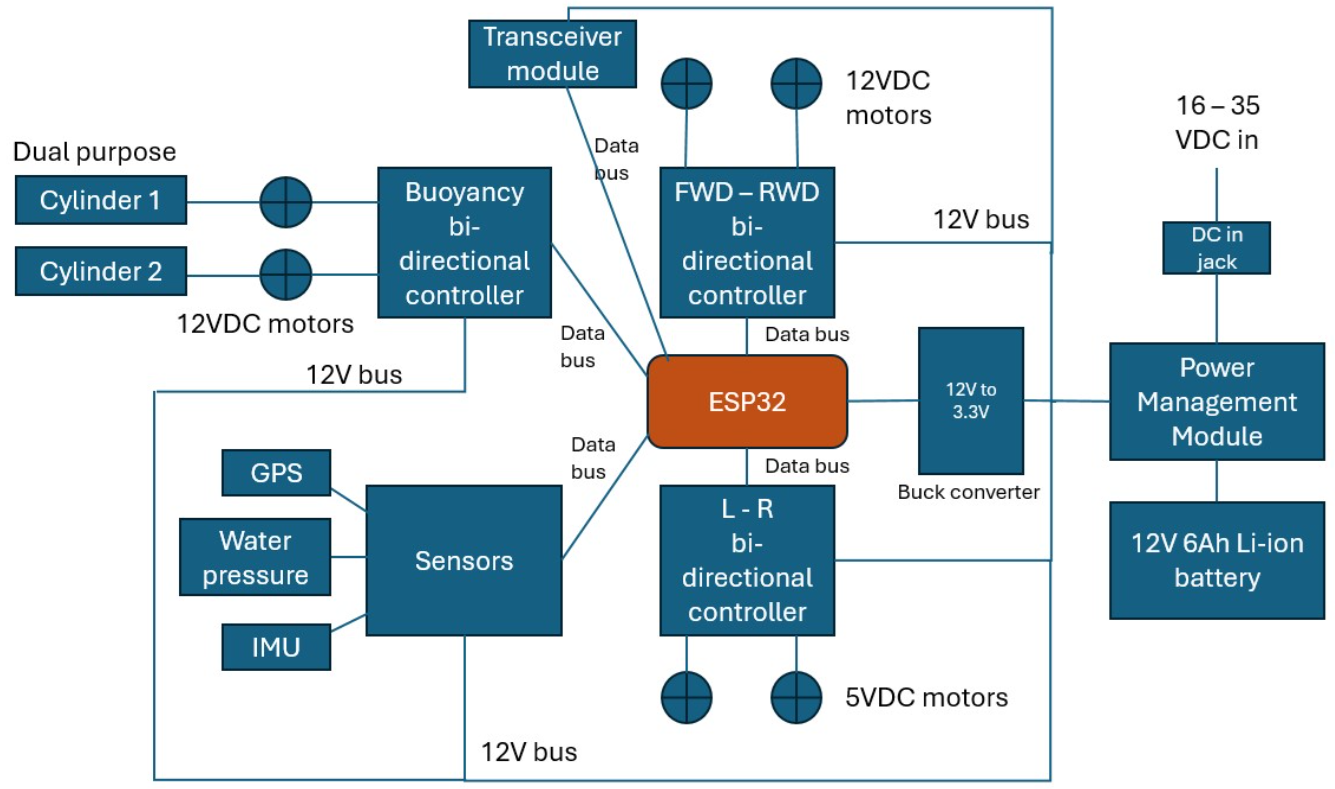}
\caption{Electronic circuit diagram}
\label{fig:circuit}
\end{figure*}

\section{Hardware and software design}

\subsection{Hardware design}

\subsubsection{External structure and propulsion}

The proposed miniature submarine uses a compact cylindrical acrylic hull that provides waterproof protection while keeping the structure lightweight and portable, as shown in Fig. \ref{fig:submarine}. The vehicle is small enough to be transported and deployed by a single operator. The transparent acrylic tube also allows easy inspection of internal components and simplifies maintenance.

For propulsion, two 12 V pump motors are mounted symmetrically at the two ends of the cylindrical hull to generate forward and reverse thrust through water jet propulsion. Each pump motor is secured using custom 3D-printed semi-cylindrical mounting brackets that conform to the curvature of the hull.

At one end of the hull, two 5 V DC pump motors are positioned laterally on the left and right sides to generate transverse water jets for yaw control. These steering pumps, together with one propulsion pump, are integrated into a curved 3D-printed mounting frame that follows the cylindrical contour of the hull and distributes mechanical loads evenly. The distributed placement of the motors lowers the center of gravity, improving roll stability and reducing the risk of flipping during operation.

\subsubsection{Buoyancy control system}

Vertical motion is achieved using two water-intake cylinders located at both ends of the hull. Each cylinder is actuated by a 12 V DC motor that drives the mechanism responsible for intaking or releasing water. By adjusting the amount of water stored in the cylinders, the submarine modifies its total mass and buoyancy, enabling ascending, descending, or neutral buoyancy control. Limit switches installed at both ends of the cylinders prevent mechanical overextension during operation.

The buoyancy mechanism follows Archimedes' principle, where the buoyant force $F_b = \rho V g$ balances the gravitational force $F_g = mg$. Since the external hull volume remains nearly constant, buoyancy is controlled by varying the internal water mass within the cylinders. Experimental validation was conducted by filling the 150 ml intake cylinders and gradually adding ballast until the vehicle reached the sinking threshold, allowing empirical calibration of the buoyancy balance.

\subsubsection{Internal electronics and control}

The onboard electronics are centered around an ESP32 microcontroller, which manages sensor acquisition, actuator control, and system coordination, as illustrated in Fig.\ref{fig:circuit}. The vehicle is powered by a 12 V lithium battery capable of supporting hours of operation under maximum load. A voltage converter steps the supply down to 5 V and 3.3 V to power sensors and logic circuits.

Because the ESP32 outputs low-current control signals, H-bridge motor drivers are used for actuation. Two L298N drivers control the propulsion pump motors and the DC motors driving the buoyancy cylinders, while a mini L298 driver controls the steering pump motors, enabling bidirectional motor operation and speed control.

\subsubsection{Sensors and communication}

To support navigation and environmental monitoring, the submarine integrates several onboard sensors, including a NEO-6M GPS module for surface positioning, an industrial water pressure sensor for depth estimation, and an IMU for orientation and motion tracking. The pressure sensor produces a linear analog voltage proportional to the applied pressure, which is digitized using the 12-bit ADC of the ESP32 to calculate water depth based on hydrostatic pressure.

Wireless communication between the submarine and the surface controller is implemented using LoRa modules. Although radio signals attenuate underwater, LoRa communication remains effective at the vehicle's shallow operating depth of 2-3 m.

\subsection{Software program}

The software architecture of the miniature submarine is designed around an ESP32 microcontroller. The program is structured into a continuous execution loop that handles synchronous sensor polling and asynchronous command processing to ensure real-time responsiveness.

\subsubsection{Data ccquisition and processing}The embedded software continuously monitors the vehicle's state by interfacing with a suite of onboard sensors:
\begin{itemize}
\item \textbf{Inertial measurement:} An I2C-based BNO055 IMU is sampled to retrieve 6-DOF kinematic data, including linear acceleration, angular velocity (gyroscope), and orientation in both Euler angles and quaternions. To estimate the vehicle's kinematics in the surge, sway, and heave axes, the software applies a numerical integration step over the linear acceleration data to calculate real-time velocity and displacement.
\item \textbf{Depth estimation:} A pressure sensor interfaced via an ADC pin measures external water pressure. The raw analog readings are scaled to kilopascals (kPa) and subsequently converted to a depth estimate (in meters) using the hydrostatic pressure equation.
\item \textbf{Surface localization:} A GPS module communicates with the ESP32 via a hardware serial UART interface, parsed using the TinyGPS++ library to obtain precise latitude and longitude coordinates when the vehicle is at the surface.
\end{itemize}

\subsubsection{Actuator control and safety mechanisms}The vehicle's motion is driven by six distinct DC motor channels controlled via bi-directional motor drivers. The software maps incoming control commands to specific high/low digital pin states to actuate the forward/backward propulsion pumps, left/right steering pumps, and the dual buoyancy control cylinders. To ensure the mechanical integrity of the buoyancy system, the software incorporates hardware-level safety interrupts. Two limit switches are connected to the ESP32 and configured with falling-edge interrupts. If a buoyancy cylinder reaches its maximum physical extension, the interrupt service routine (ISR) immediately halts the respective motor to prevent hardware damage.
\subsubsection{Communication and telemetry}For teleoperation and monitoring, the system utilizes a LoRa transceiver communicating over a 433.2 MHz frequency via the SPI bus. The communication protocol is strictly defined:
\begin{itemize}
\item \textbf{Command reception:} The ESP32 asynchronously listens for incoming LoRa packets. Received string commands are parsed into motor identification numbers and operational states (e.g., "forward", "reverse", "stop") and immediately passed to the actuator control logic.
\item \textbf{Telemetry transmission:} At a fixed 2000-millisecond interval, the software constructs a comprehensive telemetry payload. The measured data are formatted into telemetry packets and transmitted through the LoRa communication link to the surface controller.
\end{itemize}

An example received telemetry packet is shown below:

\begin{verbatim}
<<< Packet Received! RSSI: -58 | Raw:
`LAT:21.027252,LON:105.851954|ACC:0.00,-0.02,0.00|
EUL:214.00,-33.06,-6.75|GYR:-0.00,-0.00,0.00|
Q:0.28,-0.26,0.13,0.92|VEL:0.00,0.00,0.00|
DIS:193.02,42.91,0.00'
\end{verbatim}

The transmitted packets include key navigation parameters such as geographic position, acceleration, orientation, angular velocity, velocity, and displacement. Additionally, the received signal strength indicator (RSSI) is recorded to evaluate the reliability of the wireless communication link.

\begin{figure}[ht]
\includegraphics[width=\columnwidth]{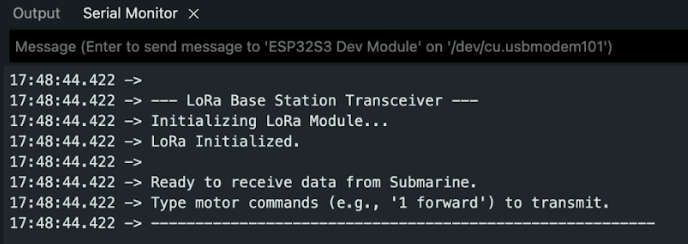}
\caption{Controller software initialization in the Arduino IDE serial monitor.}
\label{fig:software}
\end{figure}

Figure~\ref{fig:software} shows the initialization of the controller software in the Arduino IDE serial monitor. Once the system starts, the LoRa module is initialized and the controller enters standby mode, waiting to receive telemetry from the submarine and to send control commands.

\subsection{Estimated cost}

Table~\ref{tab:bom} summarizes the bill of materials (BOM) for the miniature submarine prototype. The system is constructed using low-cost, commercially available components, including sensors, microcontrollers, pumps, and mechanical parts. The total estimated hardware cost of the prototype is approximately 122.5~USD. This low cost demonstrates that the proposed design can provide a practical and accessible platform for research, education, and small-scale underwater monitoring applications.

\begin{table*}[ht]
\centering
\caption{Bill of materials (BOM) for the miniature submarine prototype}
\label{tab:bom}
\begin{tabular}{c l l c c c}
\hline
\textbf{No.} & \textbf{Item} & \textbf{Spec} & \textbf{Qty} & \textbf{Unit Price (USD)} & \textbf{Amount (USD)} \\
\hline
1 & Body & Acrylic cylinder OD 100 mm, length 750 mm & 1 & 8 & 8 \\
2 & Thruster & 12V 9W & 2 & 5 & 10 \\
3 & Steering & 5V 1W & 2 & 1 & 2 \\
4 & Battery & 12V 6Ah 10A & 1 & 15 & 15 \\
5 & DC-DC converter 5V & 5V 3A & 1 & 1.5 & 1.5 \\
6 & DC-DC converter 3.3V & 3.3V 2A & 1 & 1 & 1 \\
7 & Ballast pump & 12V 6W & 3 & 5 & 15 \\
8 & Syringe & 200 ml & 3 & 0.5 & 1.5 \\
9 & GPS receiver & U-blox Neo M10 + ceramic antenna & 1 & 15 & 15 \\
10 & Pressure sensor & Sealed BMP280 & 1 & 2 & 2 \\
11 & IMU & BNO055 9DoF & 1 & 19.5 & 19.5 \\
12 & SD card & 4GB & 1 & 2 & 2 \\
13 & SD card reader / writer & Micro SD read/write 3.3V & 1 & 0.5 & 0.5 \\
14 & 3D printed parts & Fit the components' size & 1 & 2 & 2 \\
15 & Tubing & Various for pump, syringe, sensor & 1 & 2 & 2 \\
16 & Bare PCB & 7 × 9 cm & 1 & 1 & 1 \\
17 & MCU & ESP32-S3 & 1 & 6.5 & 6.5 \\
18 & 12V dual motor driver & L298N & 2 & 1.5 & 3 \\
19 & 5V dual motor driver & L298 mini & 1 & 1.5 & 1.5 \\
20 & Water resistant connectors & ID 6 mm, OD 12 mm & 6 & 0.2 & 1.2 \\
21 & Water resistant cables & Various for thruster and steering motors & 1 & 2 & 2 \\
22 & Charging diode & 20A Schottky diode for watertight electrodes & 1 & 0.5 & 0.5 \\
23 & LoRa transceiver & SX1278 433 MHz & 1 & 4.8 & 4.8 \\
24 & Ballast pump and syringe frame & M4 stainless steel nuts and bars & 1 & 2 & 2 \\
25 & Sealant & Silicon-based sealant & 1 & 3 & 3 \\
\hline
\multicolumn{5}{r}{\textbf{Total Cost}} & \textbf{122.5 USD} \\
\hline
\end{tabular}
\end{table*}

\section{Experimental results}

\begin{figure}[ht]
\includegraphics[width=\columnwidth]{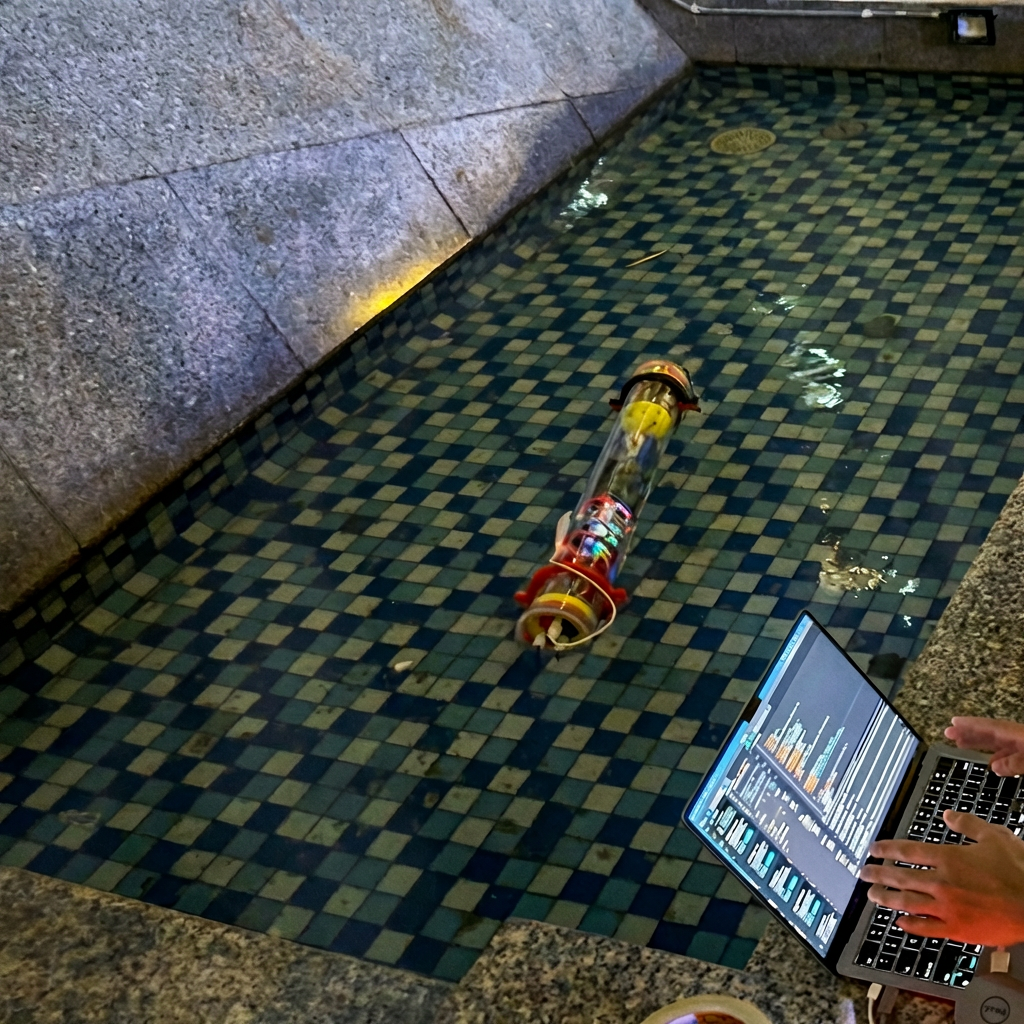}
\caption{Experimental setup in a pool.}
\label{fig:pool}
\end{figure}

A series of experiments were conducted to evaluate the propulsion, steering, and buoyancy control performance of the proposed miniature submarine. Tests included controlled experiments in a small pool (Fig.\ref{fig:pool}) and field trials in a lake (Figs.\ref{fig:lake}). Key metrics such as attitude stability, steering response, depth regulation, and navigation capability were analyzed.

\begin{figure}[ht]
\includegraphics[width=\columnwidth]{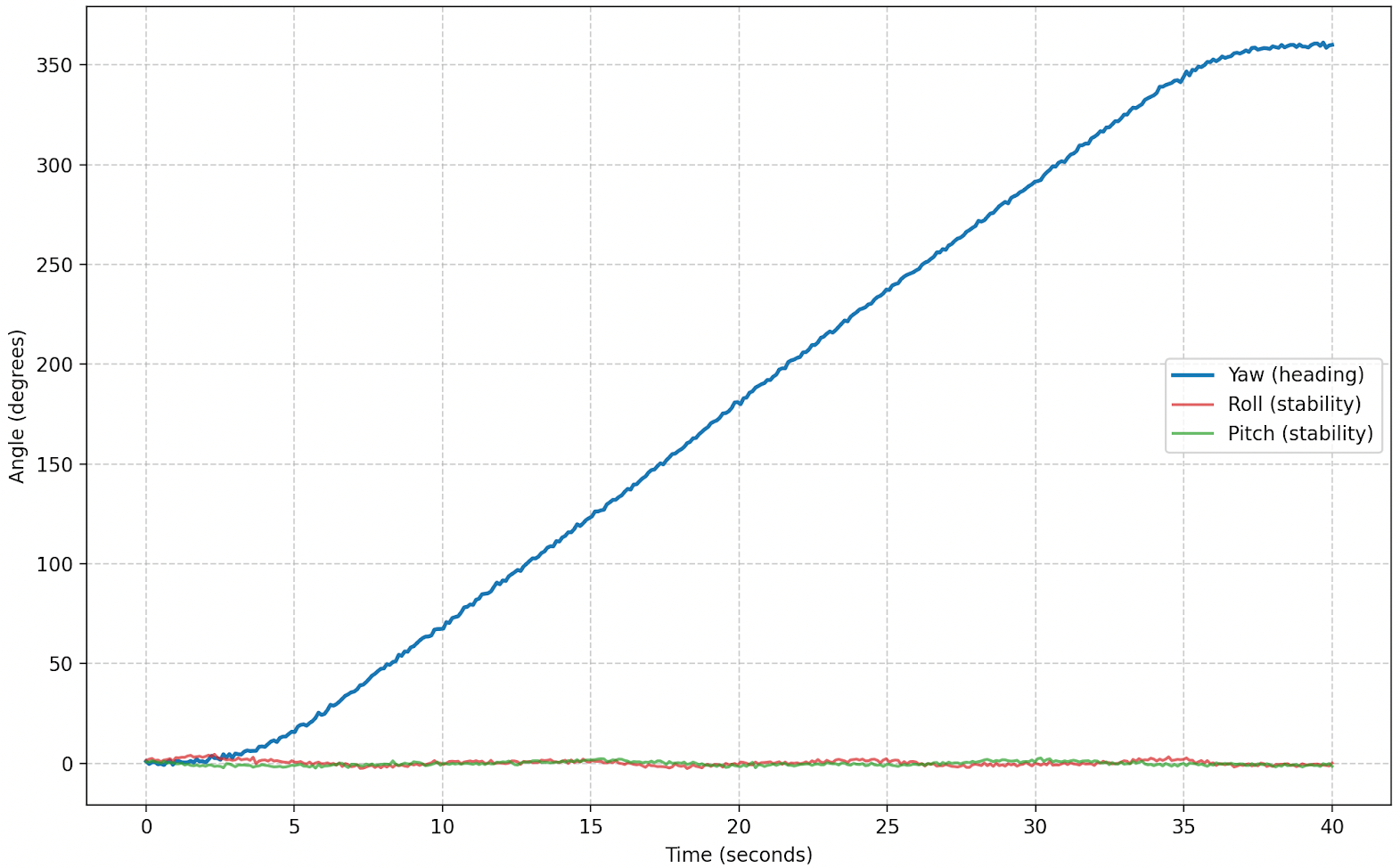}
\caption{Attitude stability during a $360^0$ yaw turn.}
\label{fig:att}
\end{figure}

\subsection{Pool experiments}

\subsubsection{Attitude stability}

To evaluate dynamic stability, the submarine performed a full $360^{\circ}$ yaw rotation at the water surface over a 40\,s interval while heading, roll, and pitch angles were recorded using the onboard IMU (Fig.\ref{fig:att}). The heading increased smoothly from $0^{\circ}$ to approximately $360^{\circ}$, corresponding to an average yaw rate of about $9^{\circ}$/s. During the maneuver, roll and pitch remained confined to small oscillations around $0^{\circ}$. The maximum roll deviation was approximately $\pm2^{\circ}$, while the maximum pitch deviation remained within $\pm1.5^{\circ}$. No significant drift or coupling between axes was observed. These results confirm that the clustered thruster configuration and low center-of-gravity design maintain stable attitude during rotational motion.

\begin{figure}[ht]
\includegraphics[width=\columnwidth]{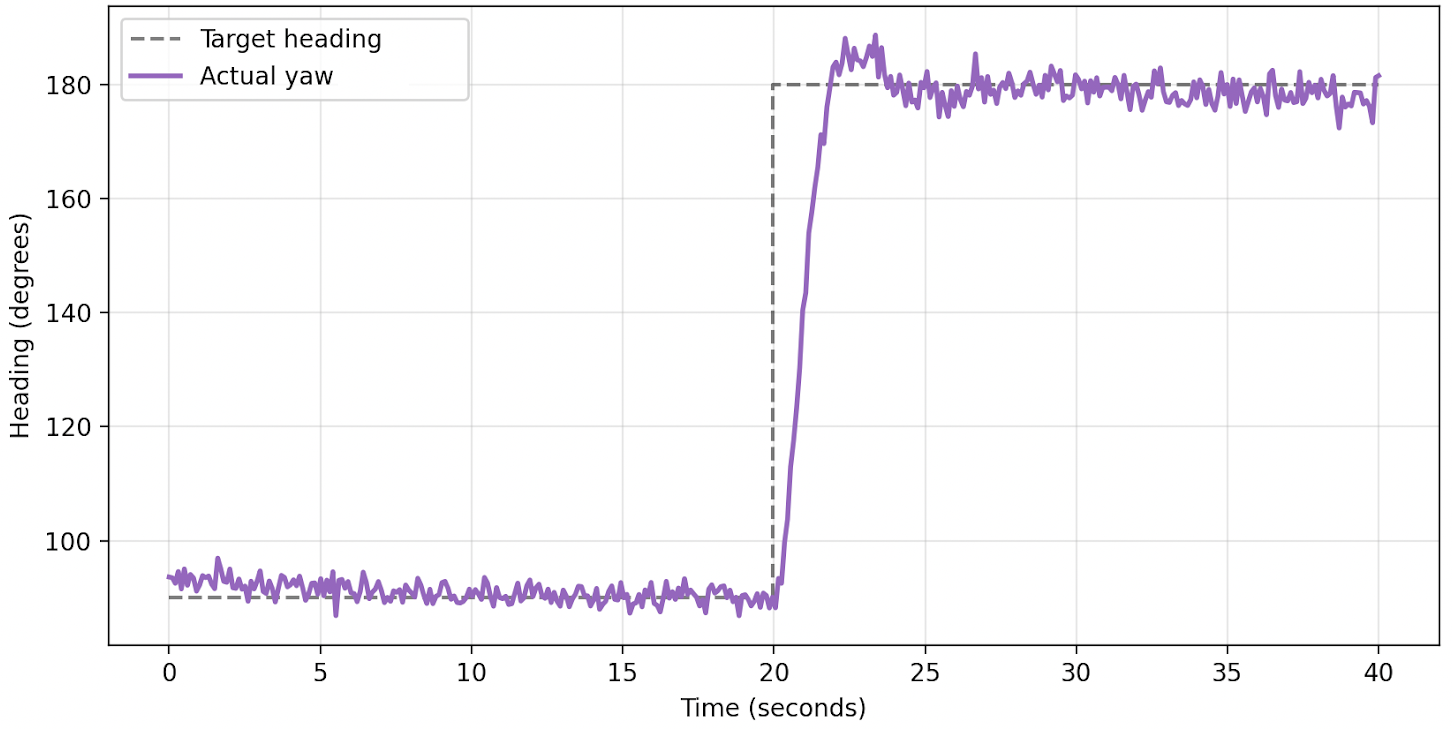}
\caption{Differential drive steering response.}
\label{fig:steer}
\end{figure}

\subsubsection{Differential drive steering response}

The responsiveness of the steering control algorithm was evaluated by changing the desired heading from $90^{\circ}$ to $180^{\circ}$ at $t=20$\,s (Fig.\ref{fig:steer}). The measured yaw tracked the commanded heading with a transient response characterized by a rise time of approximately $2$\,s. A small overshoot of about $5^{\circ}$ was observed before the heading converged to the new setpoint. The system reached steady state within approximately $4$--$5$\,s with a steady-state error below $2^{\circ}$. These results demonstrate that differential thrust provides effective directional control for the vehicle.

\begin{figure}[ht]
\includegraphics[width=0.9\columnwidth]{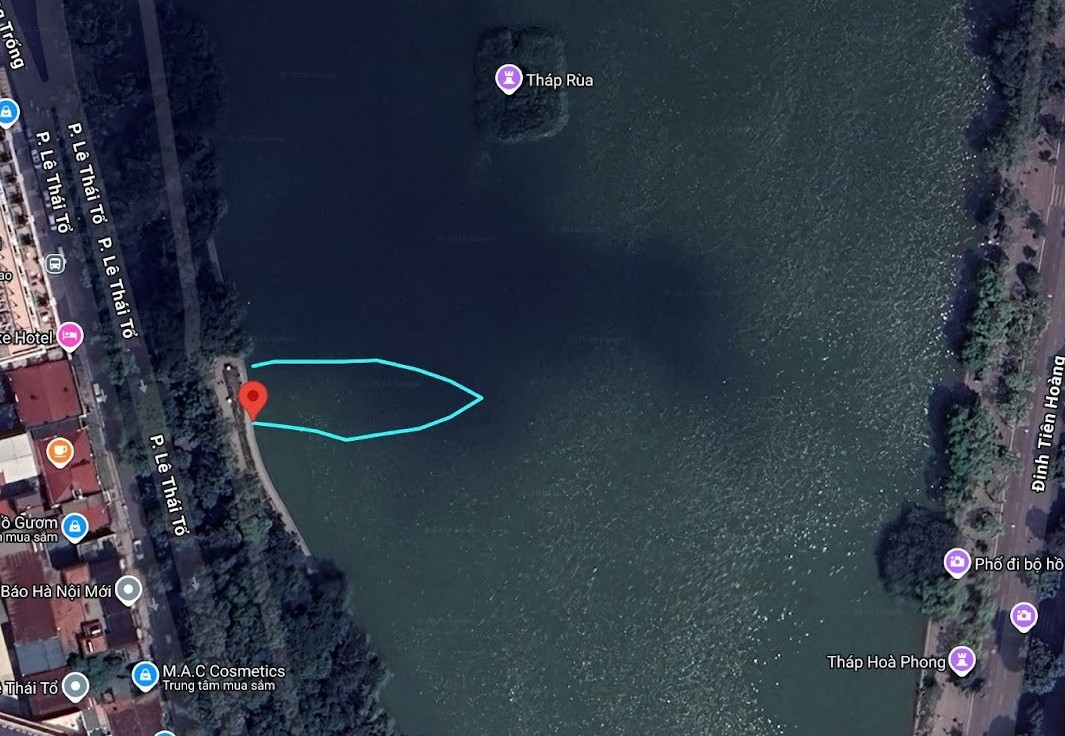}
\caption{GPS track of a water navigation session in a lake.}
\label{fig:lake}
\end{figure}

\begin{figure}[ht]
\includegraphics[width=\columnwidth]{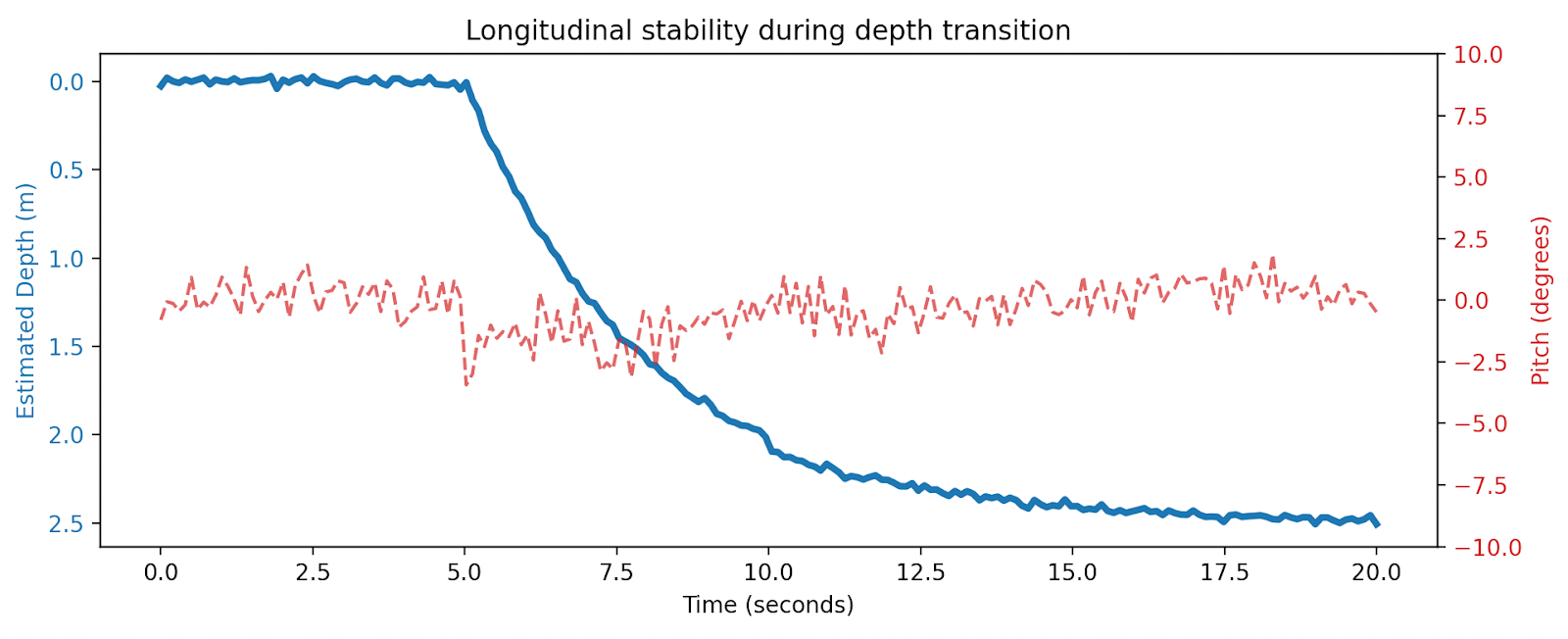}
\caption{Longitudinal stability during depth transition.}
\label{fig:depth}
\end{figure}

\subsection{Field experiments}

\subsubsection{Surface navigation}

Field trials were conducted in a lake to evaluate real-world navigation performance. Figure \ref{fig:lake} shows the GPS trajectory recorded during a navigation session. The submarine successfully traveled approximately 35-40\,m from the shoreline to the sampling location and come back while maintaining stable communication with the surface station via the LoRa link.

\subsubsection{Depth control and longitudinal stability}

The buoyancy control system was validated by commanding the submarine to descend from the surface to a target depth of approximately $2.5$\,m (Fig.\ref{fig:depth}). The descent started at $t \approx 5$\,s and reached the target depth within about $8$-$10$\,s. The steady-state depth error remained within approximately $\pm0.1$\,m, with small fluctuations due to pressure sensor noise. During the transition, the pitch angle remained bounded within approximately $\pm3^{\circ}$, indicating stable longitudinal balance and effective compensation for internal mass redistribution. The results confirm that the ballast pump system enables smooth and controlled depth regulation without excessive nose-down motion.

\begin{figure}[ht]
\includegraphics[width=\columnwidth]{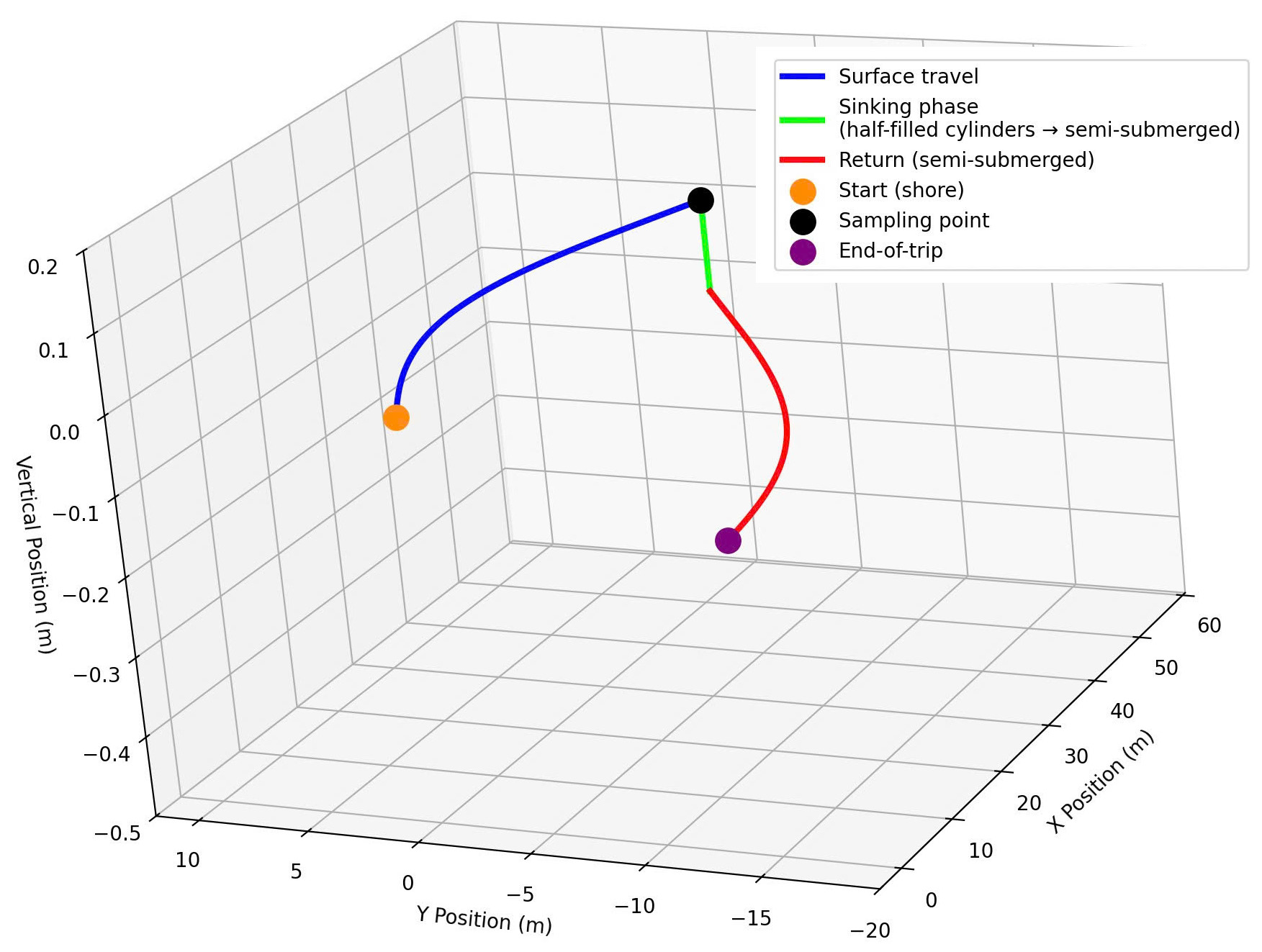}
\caption{A water sampling session.}
\label{fig:sample}
\end{figure}

\subsubsection{Water sampling mission}

A complete sampling mission was conducted in a lake, as shown in Fig.~\ref{fig:sample}. The submarine first traveled along the surface to the sampling point and then partially submerged to the desired depth. Water samples were collected using the dual-purpose cylinders that serve both as buoyancy control units and sampling chambers. By actuating the cylinders, surrounding water is drawn into the chambers during the sampling phase. After sample collection, the submarine returned toward the shore. The trajectory shows a horizontal travel distance of approximately 40 m and a vertical descent of about $0.13$ m during the sampling phase. The experiment demonstrates the feasibility of using the integrated buoyancy and sampling cylinders for environmental monitoring tasks.

Overall, the experimental results confirm that the proposed miniature submarine can maintain stable attitude, perform controlled steering maneuvers, regulate depth reliably, and execute practical water sampling missions in real-world environments.

\section{Conclusion}

This paper presented the design and experimental validation of a bio-inspired miniature submarine for low-cost water quality monitoring. Inspired by squid jet propulsion, the proposed platform employs pump-driven propulsion and steering together with a pump-based buoyancy system that enables both depth regulation and water sampling. The system integrates low-cost components, including an ESP32 microcontroller, IMU, pressure sensor, GPS, and LoRa communication. The total hardware cost is approximately \$122.5. Experimental results from pool and field tests demonstrated stable motion and reliable control, with roll and pitch deviations within $\pm2^\circ$ and $\pm1.5^\circ$, a steering response with about 2\,s rise time and 5\,s settling time, and depth regulation at 2.5\,m with a steady-state error within $\pm0.1$\,m. These results confirm that the proposed platform provides an affordable and practical solution for small-scale aquatic environmental monitoring. Future work will focus on integrating additional water quality sensors and improving autonomous navigation capabilities.

\balance
\bibliographystyle{ieeetr}  
\bibliography{bibtex}

\begin{thebibliography}{10}

\bibitem{chapman2022role}
D.~V. Chapman and T.~Sullivan, ``The role of water quality monitoring in the sustainable use of ambient waters,'' {\em One Earth}, vol.~5, no.~2, pp.~132--137, 2022.

\bibitem{rand2022human}
J.~M. Rand, M.~O. Nanko, M.~B. Lykkegaard, D.~Wain, W.~King, L.~D. Bryant, and A.~Hunter, ``The human factor: Weather bias in manual lake water quality monitoring,'' {\em Limnology and Oceanography: Methods}, vol.~20, no.~5, pp.~288--303, 2022.

\bibitem{lopez2023wireless}
G.~A. Lopez-Ramirez and A.~Aragon-Zavala, ``Wireless sensor networks for water quality monitoring: a comprehensive review,'' {\em IEEE access}, vol.~11, pp.~95120--95142, 2023.

\bibitem{petillot2019underwater}
Y.~R. Petillot, G.~Antonelli, G.~Casalino, and F.~Ferreira, ``Underwater robots: From remotely operated vehicles to intervention-autonomous underwater vehicles,'' {\em IEEE Robotics \& Automation Magazine}, vol.~26, no.~2, pp.~94--101, 2019.

\bibitem{cai2023cooperative}
W.~Cai, Z.~Liu, M.~Zhang, and C.~Wang, ``Cooperative artificial intelligence for underwater robotic swarm,'' {\em Robotics and Autonomous Systems}, vol.~164, p.~104410, 2023.

\bibitem{1240276}
F.~Nauert and P.~Kampmann, ``Inspection and maintenance of industrial infrastructure with autonomous underwater robots,'' {\em Frontiers in Robotics and AI}, vol.~Volume 10 - 2023, 2023.

\bibitem{costa2018design}
D.~Costa, G.~Palmieri, M.-C. Palpacelli, L.~Panebianco, and D.~Scaradozzi, ``Design of a bio-inspired autonomous underwater robot,'' {\em Journal of Intelligent \& Robotic Systems}, vol.~91, no.~2, pp.~181--192, 2018.

\bibitem{park2014design}
Y.-J. Park, T.~M. Huh, D.~Park, and K.-J. Cho, ``Design of a variable-stiffness flapping mechanism for maximizing the thrust of a bio-inspired underwater robot,'' {\em Bioinspiration \& biomimetics}, vol.~9, no.~3, p.~036002, 2014.

\bibitem{xiong2023bio}
X.~Xiong, H.~Xu, S.~Chen, H.~Wang, C.~You, and Y.~Wu, ``A bio-inspired underwater robot inspired by jellyfish,'' in {\em Fluid Power Systems Technology}, vol.~87431, p.~V001T01A021, American Society of Mechanical Engineers, 2023.

\bibitem{neira2021review}
J.~Neira, C.~Sequeiros, R.~Huamani, E.~Machaca, P.~Fonseca, and W.~Nina, ``Review on unmanned underwater robotics, structure designs, materials, sensors, actuators, and navigation control,'' {\em Journal of Robotics}, vol.~2021, no.~1, p.~5542920, 2021.

\bibitem{10477402}
K.~Hasan, S.~Ahmad, A.~F. Liaf, M.~Karimi, T.~Ahmed, M.~A. Shawon, and S.~Mekhilef, ``Oceanic challenges to technological solutions: A review of autonomous underwater vehicle path technologies in biomimicry, control, navigation, and sensing,'' {\em IEEE Access}, vol.~12, pp.~46202--46231, 2024.

\bibitem{tolkoff1999robotics}
S.~W. Tolkoff, {\em Robotics and power measurements of the RoboTuna}.
\newblock PhD thesis, Massachusetts Institute of Technology, 1999.

\bibitem{li2022comprehensive}
Y.~Li, Y.~Xu, Z.~Wu, L.~Ma, M.~Guo, Z.~Li, and Y.~Li, ``A comprehensive review on fish-inspired robots,'' {\em International Journal of Advanced Robotic Systems}, vol.~19, no.~3, p.~17298806221103707, 2022.

\bibitem{dabiri2009optimal}
J.~O. Dabiri, ``Optimal vortex formation as a unifying principle in biological propulsion,'' {\em Annual review of fluid mechanics}, vol.~41, no.~1, pp.~17--33, 2009.

\bibitem{raj2016fish}
A.~Raj and A.~Thakur, ``Fish-inspired robots: design, sensing, actuation, and autonomy—a review of research,'' {\em Bioinspiration \& biomimetics}, vol.~11, no.~3, p.~031001, 2016.

\bibitem{yan2024recent}
S.~Yan, Z.~Wu, J.~Wang, Y.~Feng, L.~Yu, J.~Yu, and M.~Tan, ``Recent advances in design, sensing, and autonomy of biomimetic robotic fish: A review,'' {\em IEEE/ASME Transactions on Mechatronics}, 2024.

\bibitem{wang2020development}
R.~Wang, S.~Wang, Y.~Wang, L.~Cheng, and M.~Tan, ``Development and motion control of biomimetic underwater robots: A survey,'' {\em IEEE Transactions on Systems, Man, and Cybernetics: Systems}, vol.~52, no.~2, pp.~833--844, 2020.

\bibitem{4623834}
M.~Krieg and K.~Mohseni, ``Thrust characterization of a bioinspired vortex ring thruster for locomotion of underwater robots,'' {\em IEEE Journal of Oceanic Engineering}, vol.~33, no.~2, pp.~123--132, 2008.

\bibitem{shen2017biomimetic}
Z.~Shen, J.~Na, and Z.~Wang, ``A biomimetic underwater soft robot inspired by cephalopod mollusc,'' {\em IEEE Robotics and Automation Letters}, vol.~2, no.~4, pp.~2217--2223, 2017.

\bibitem{marchese2014autonomous}
A.~D. Marchese, C.~D. Onal, and D.~Rus, ``Autonomous soft robotic fish capable of escape maneuvers using fluidic elastomer actuators,'' {\em Soft robotics}, vol.~1, no.~1, pp.~75--87, 2014.

\bibitem{bao2023review}
P.~Bao, L.~Shi, L.~Duan, S.~Guo, and Z.~Li, ``A review: From aquatic lives locomotion to bio-inspired robot mechanical designations,'' {\em Journal of Bionic Engineering}, vol.~20, no.~6, pp.~2487--2511, 2023.

\bibitem{manjakkal2021connected}
L.~Manjakkal, S.~Mitra, Y.~R. Petillot, J.~Shutler, E.~M. Scott, M.~Willander, and R.~Dahiya, ``Connected sensors, innovative sensor deployment, and intelligent data analysis for online water quality monitoring,'' {\em IEEE Internet of Things Journal}, vol.~8, no.~18, pp.~13805--13824, 2021.

\bibitem{dunbabin2009autonomous}
M.~Dunbabin, A.~Grinham, and J.~Udy, ``An autonomous surface vehicle for water quality monitoring,'' in {\em Australasian conference on robotics and automation (ACRA)}, pp.~2--4, 2009.

\bibitem{adu2017water}
K.~S. Adu-Manu, C.~Tapparello, W.~Heinzelman, F.~A. Katsriku, and J.-D. Abdulai, ``Water quality monitoring using wireless sensor networks: Current trends and future research directions,'' {\em ACM Transactions on Sensor Networks (TOSN)}, vol.~13, no.~1, pp.~1--41, 2017.

\bibitem{madeo2020low}
D.~Madeo, A.~Pozzebon, C.~Mocenni, and D.~Bertoni, ``A low-cost unmanned surface vehicle for pervasive water quality monitoring,'' {\em IEEE Transactions on Instrumentation and Measurement}, vol.~69, no.~4, pp.~1433--1444, 2020.

\bibitem{LI2022109056}
Z.~Li, X.~Chao, I.~Hameed, J.~Li, W.~Zhao, and X.~Jing, ``Biomimetic omnidirectional multi-tail underwater robot,'' {\em Mechanical Systems and Signal Processing}, vol.~173, p.~109056, 2022.

\bibitem{202406956}
D.~Wang, F.~Zhang, S.~Zhang, D.~Liu, J.~Li, W.~Chen, J.~Deng, and Y.~Liu, ``Miniature modular reconfigurable underwater robot based on synthetic jet,'' {\em Advanced Science}, vol.~11, no.~39, p.~2406956, 2024.

\end{thebibliography}

\end{document}